\documentclass[10pt,letterpaper]{article}

\usepackage{cvpr}
\usepackage{times}
\usepackage{epsfig}
\usepackage{graphicx}
\usepackage{amsmath}
\usepackage{amssymb}
\usepackage{algorithm2e}
\usepackage{tabu}
\usepackage{tabularx}
\usepackage{multirow}
\usepackage{floatrow}
\usepackage[breaklinks=true,bookmarks=false]{hyperref}

\cvprfinalcopy 

\begin{document}

\title{One - Click Annotation with Guided Hierarchical Object Detection}

\author{ {Adithya Subramanian,} {Anbumani Subramanian}\\
Intel\\
Bangalore, India\\
{\tt\small adithya.subramanian@intel.com}
\\
{\tt\small anbumani.subramanian@intel.com}
}

\maketitle

\begin{abstract}
The increase in data collection has made data annotation an interesting and valuable task in the contemporary world. This paper presents a new methodology for quickly annotating data using click-supervision and hierarchical object detection. The proposed work is semi-automatic in nature where the task of annotations is split between the human and a neural network. We show that our improved method of annotation reduces the time, cost and mental stress on a human annotator. The research also highlights how our method performs better than the current approach in different circumstances such as variation in number of objects, object size and different datasets. Our approach also proposes a new method of using object detectors making it suitable for data annotation task. The experiment conducted on PASCAL VOC dataset revealed that annotation created from our approach achieves a mAP of 0.995 and a recall of 0.903. The Our Approach has shown an overall improvement by 8.5\%, 18.6\% in mean average precision and recall score for KITTI and 69.6\%, 36\% for CITYSCAPES dataset. The proposed framework is 3-4 times faster as compared to the standard annotation method.

\end{abstract}

\section{Introduction}

Annotated data is an extremely valuable asset in both academia and industry. The availability of data has provided data annotation task a greater importance in the society. A lot of deep learning research has been focused on improving the generalizability of the deep neural networks with only a small set of training samples which is known as few shot learning \cite{snell2017prototypical,garcia2017few,hilliard2018few,sung2017learning}. In contrary to this type of research a little attention has been shown in improving the annotation process to make more data available for allowing models to generalize well.
\\\\
The current annotation strategy for object detection involves clicking the object on the left-top and right bottom of the image but this task puts the user into heavy mental stress and also a huge amount of time is consumed in the process of finding an extremely tight bounding box. The same has been proved in multiple research papers \cite{konyushkova2017learning,papadopoulos2017training,papadopoulos2017extreme}. To avoid the mental stress as well as the time consumption, we propose a semi - automatic approach which combines best of the both worlds i.e. the accuracy of human eye and the speed of neural networks.
\\\\
The current object detectors cannot be used for the purpose of annotating data as they have a low recall score and mean average precision score. This leads to unreliable results where the object detector might have miss-classified an object or it might have made a completely wrong prediction both in terms of classification and localization. The low recall score also suffers from the same issue but now the problem becomes severe as the detector classifies an object as a background. When a new network is trained with these annotation then it won't able to generalize well for these incorrectly classified, incorrectly localized and missed out object category in the annotations.  
\\\\
Our approach is rather robust to these issues. The proposed framework being semi-automatic it only acquires partial annotation from the annotator by making them click on the object centers. The framework at the same time simultaneously predicts the intermediate detections from the object detector. These detections are further refined from these human annotated object centers removing the incorrect classifications as well as the localizations. These object centers are further used by the detector to create object proposals when it misses to predict objects at this object center. The created object proposals are further used by the detector as an input to the network for detecting the objects. This process is iterated hierarchically until the object is detected.
\\\\
The further sections are ordered such that Section 2 highlights the related work, Section 3 describes the proposed work. Experimental results are discussed in Section 4 and conclusion for the work is derived at Section 5. The references for this work is listed after the conclusion.

\section{Related work}
\label{sec:related_work}

The research in the field of deep learning has focused on using unlabelled or partially labelled data by developing model in semi-supervised, unsupervised or weakly-supervised learning paradigm to reduce the dependence of the model over annotated data. In the semi - supervised learning paradigm Tan et al. \cite{tang2017visual} has proposed a novel algorithm for large scale semi-supervised object detection by imbibing knowledge from visual and semantic cues. Rhee et al \cite{rhee2017active} has developed an object detector in semi - supervised learning paradigm which is initially trained on a set of perfectly labelled examples and then it uses active learning to batch imperfect and unlabelled samples. The weakly supervised learning has also been in spotlight lately. Li et al. \cite{li2016weakly} has worked on weakly supervised learning by making use of progressive domain adaptation to solve the problem of model initialization and local minima convergence which is a common issue in weakly - supervised learning paradigm. Zhang et al. \cite{zhang2017bridging} has proposed a new state-of-the-art weakly supervised model which combines saliency detection and weakly supervised object detection based on self-paced curriculum learning. There has also been research work where models have been created which can work in both weakly - supervised and semi-supervised paradigm such as \cite{yan2017weakly}.
\\\\
The fatal problem with all these previous approaches is again the requirement of data to generalize well. The semi-supervised as well as the weakly-supervised approaches barely the match performance of a fully supervised object detector such as Faster - RCNN \cite{ren2015faster}, Single Shot Multi-Box Detector \cite{liu2016ssd}, YOLO9000 \cite{redmon2017yolo9000} and RetinaNet \cite{lin2017focal}. The availability of data thus proves to be the easiest method to attain state-of-the-art performance in deep learning based object detection task.
\\\\
The researchers realized this fact and started to develop tools as well as automated algorithms to annotate data efficiently to reduce the human effort but the progress in this direction is scant. Bianco et al \cite{bianco2015interactive} has developed a tool which uses algorithms like linear interpolation, template matching and as well as supervised object detector depending on mode of operation which can be manual, semi - automatic or fully automatic aiding the annotator to speed-up the annotation, allowing the deep networks to learn from a considerably large annotated data. Bouquet et al. \cite{fagot2014fast} has attempted to annotate video's by propagating the annotation throughout the frames using an offline tracker followed by dynamic programming and distance transformation to penalize to the displacement between frames. Konyushkova et al \cite{konyushkova2017learning} has shown a different perspective of human - computer interaction for data-annotation by choosing the best sequence of actions to annotate images in least amount of time. This is learnt based on the previous experience which is achieved by using Q- learning to learn an approximate optimal policy. Similar interactive annotation methods has also been explored in semantic segmentation annotation task in the following works \cite{boykov2001interactive,rother2004grabcut,castrejon2017annotating,dutt2013predicting,jain2016click,shankar2015video}.
\\\\
On other hand recent research by Papadopoulos et al. \cite{papadopoulos2017training} explores using object centers as a supervision to Multiple Instance Learning frameworks for visual detection task which can make use of the data available in the Internet. Papadopoulos et al. \cite{papadopoulos2017extreme} has also worked on creating bounding boxes using four clicks on the extreme left, top, right and bottom to more intuitively annotate the data which results in a 7 times faster annotation method.

\section{Proposed Work}

The proposed work can be divided into 2 sub-sections where the first section discuss the method followed to attain the object centers which occupies minimal amount of user interaction but captures maximum information. This section also provides details on the complete work-flow of each and every step taken by the annotator to annotate the data. The second section explains the methodology followed to achieve state-of-the-art performance in case of click guided object detection task.

\subsection{Mechanism to capture annotations}
This section explains the steps followed to acquire one-click annotation i.e the object centers of the objects in the image. The first step is to display the input image to annotator where annotator selects the class to be annotated as shown in Fig 1. Once the class is selected, user can then click on the center point of the objects belonging to the selected class as shown in Fig 2. This process stores the object centers along with the associated class information and an instant feedback is provided by a red dot making user aware of the clicks made.The user then changes the class which is to be annotated. These steps are repeated continuously until all the object centers are captured. Once process is finished the annotation information is passed onto the network and the annotation results are provided which can be used by annotator to improve the clicking accuracy as shown in Fig 3. \\

\begin{figure}[h]
    \centering
    \graphicspath{ {./images/} }
    \includegraphics[scale=0.35]{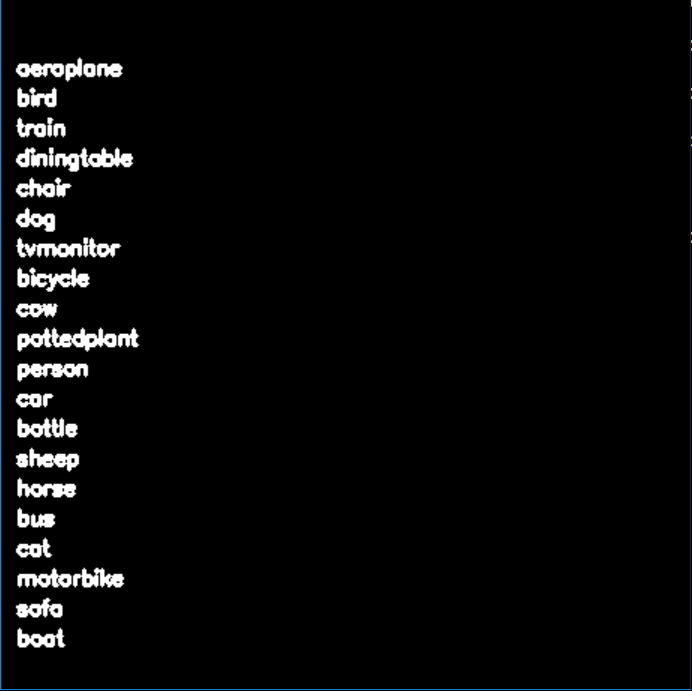}
    \caption{The class selection page}
    \label{fig:Fig 1}
\end{figure}
\begin{figure}[h]
    \centering
    \graphicspath{ {./images/} }
    \includegraphics[scale=0.45]{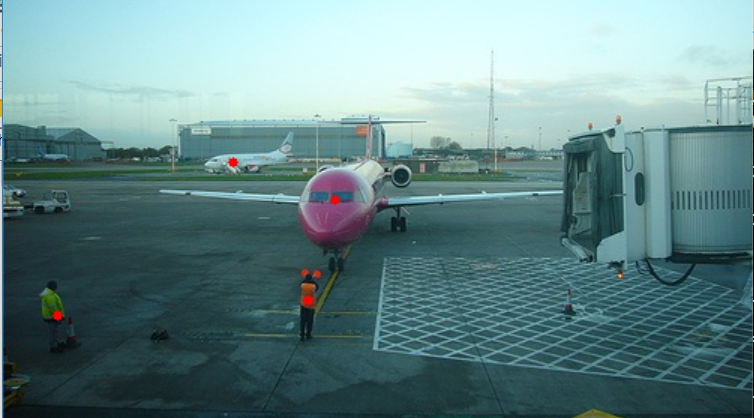}
    \caption{Instant click feedback}
    \label{fig:Fig 2}
\end{figure}

\begin{figure}
    \centering
    \graphicspath{ {./images/} }
    \includegraphics[scale=0.5]{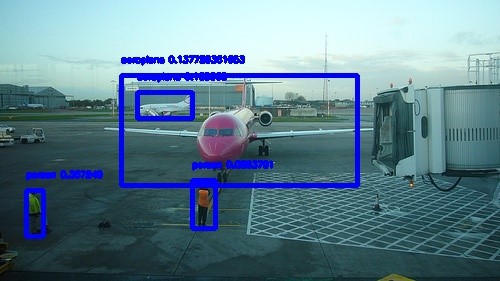}
    \caption{The bounding box results with probability of the object}
    \label{fig:Fig 3}
\end{figure}

\subsection{Hierarchical object Detection}
Hierarchical object detection consists of a base detector which is trained on a dataset having the same labels as that of the data which is to be annotated and the base detector used in the experiment is YOLO9000 which was trained to a fair amount of loss of 1.3332. The working pipeline of the detector can be broken down into a sequence of steps as seen in Fig 4,5 and the pseudo code for the same can be seen in Algorithm 1.
\begin{figure}
    \centering
    \graphicspath{ {./images/} }
    \includegraphics[scale=0.1,angle = 0]{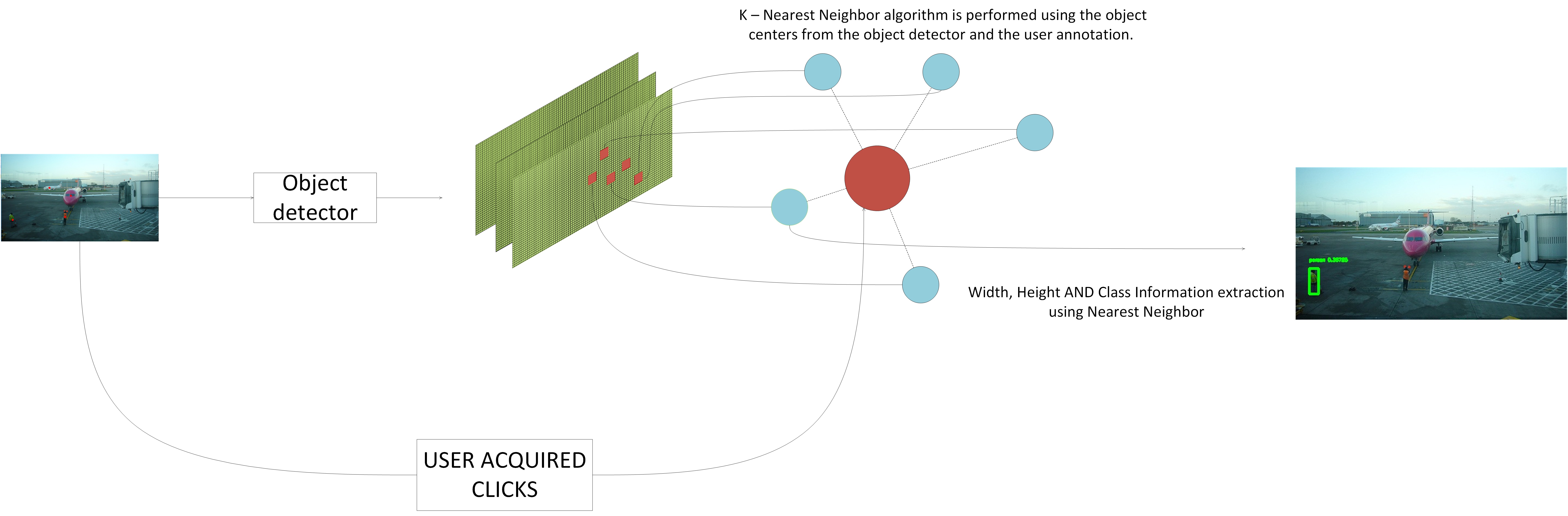}
    \caption{Improving results from standard detector using one-click annotation}
    \label{fig:Fig 4}
\end{figure}

\begin{figure}
    \graphicspath{ {./images/} }
    \includegraphics[scale=0.10,angle = 0]{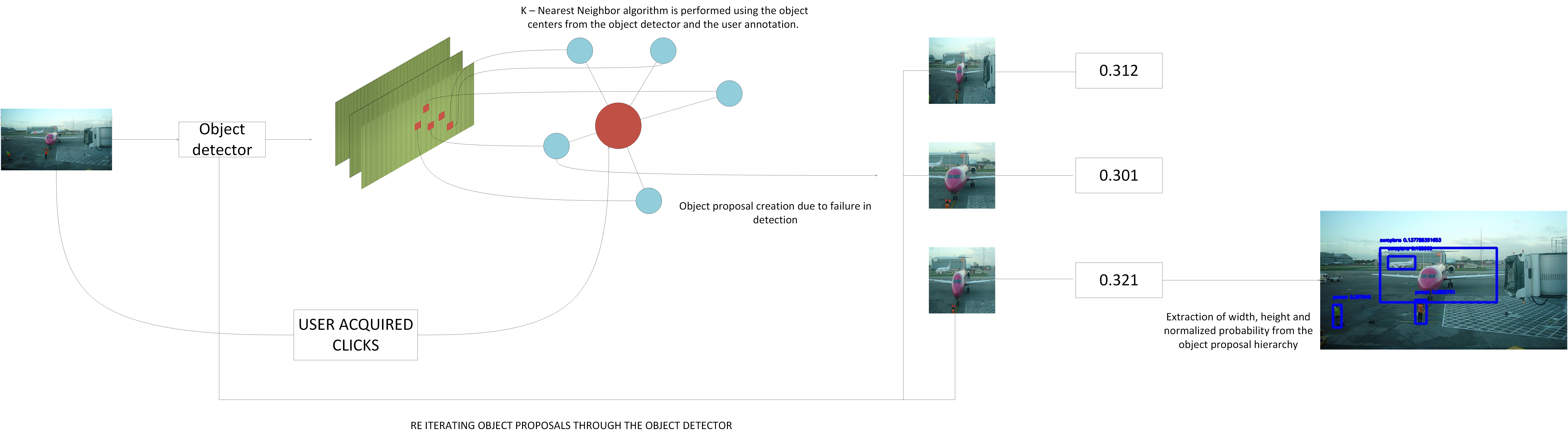}
    \caption{Improving results from standard detector using guided hierarchical object detection}
    \label{fig:Fig 5}
\end{figure}

\begin{algorithm}
\small{
\SetAlgoLined
\KwResult{Guided Hierarchical object detection}
\begin{enumerate}
        \itemsep0em 
        \item initialize X to an empty list
        \item initialize Y to an empty list
        \item initialize class to an empty list\\
\end{enumerate}
 \While{all objects are not annotated}{
 \begin{enumerate}
     \itemsep0em 
     \item Click on the object $o^{i}$ to be annotated in palette window.
     \item $class \leftarrow o^{i}$
     \item Click on the selected object's center $x_{c}$ , $y_{c}$  in the image window.
     \item $X \leftarrow x_{c}$
     \item $Y \leftarrow y_{c}$
 \end{enumerate}
}

\begin{enumerate}
\setcounter{enumi}{3}
\item N $\leftarrow$ number of network predictions
\item initialize W to an empty list
\item initialize H to an empty list
\item K $\leftarrow$ number of anchor boxes
\item T $\leftarrow$ length of the object proposal tree
\item $h \leftarrow 0$
\item $i \leftarrow 0$
\end{enumerate}
\While{i $<$ number of clicks}{
Find the K nearest neighbors i.e. the network detected object centers of the $i^{th}$ object center.\\
\eIf{if the detected neighbors are belonging to same class and closer than the threshold distance}{
\begin{enumerate}
    \itemsep0em
    \item choose the closest point.
    \item among them choose the one with highest probability.
    \item W $\leftarrow$ width of the closest neighbor
    \item H $\leftarrow$ height of the closest neighbor
    \item return the centers, width, height and probability of the bounding box
\end{enumerate}
}{
S1 :
\eIf{$h < T$}
{

$h \leftarrow h + 1$ \\ 
\lFor{for all the missed out objects centers}{
\begin{enumerate}
    \item Extract the width and height of the anchors located at these object centers.
    \item The width, height along with the object centers are used to create object proposals.
    \item apply hierarchical object detection to object proposals.\\
    \If{if there are no detection}{ goto : S1}
    \item 
    \begin{sloppypar}
    Choose the bounding box which has the highest probability among all the object proposals and \\
     return the object centers,width, height and the probability of the object in it to the higher level.
     \end{sloppypar}
\end{enumerate}}
}
{
1. return empty box
}
}
}

\caption{Guided hierarchical object detection}}
\end{algorithm}
The framework uses the click data collected from the annotator consisting of object centers and the class it belongs. These object centers and it's associated class information are used to validate the results obtained from the standard object detector and depending on the resulting accuracy of the standard object detector hierarchical object detection is performed. If the standard object detector is successful in detecting all the objects in the image then the predicted object centers are replaced with that of the annotated object centers and if the object is classified wrongly then its class information is replaced with the acquired class information. Similarly if any background information is classified as an object it is also filtered out with the help of these human annotation. If the object detector fails to detect objects in the image then hierarchical object detection comes into play, this way the time consumption of the annotation process is reduced. Hierarchical object detection plays an important role in improving the recall score of the model as the human annotation alone can only improve the mean average precision but not the recall score.
\\\\
The framework first detects the location where the standard object detector has failed to detect an object by comparing the annotation from the human and the model. Object proposals are created at these locations using the width,height information associated with the anchors boxes located in the same grid. These object proposals are further fed into the object detector for object detection. The process of generating object proposals is applied continuously until all the missed out objects are detected. This makes the task of object detection hierarchical in nature. The detection results among all the object proposals at any level is chosen based on the probability of these detection and these results are propagated back to the higher level of the hierarchy once the best among them is chosen. The probability score of the detection transitioning from a lower to higher level is multiplied by the confidence value at higher level. The intuition behind this step is make sure that the annotator knows the difficulty faced by the detector to annotate the object which can used for post-processing of the coarse annotations. Any particular branch in the object proposal tree is expanded only when the resulting probability of the detection from the particular branch will be higher than the neighbouring branches which can be seen in Fig 6, this helps in reducing the computation time and memory consumption removing the dependency of the framework over the high compute power devices.

\begin{figure}
    \centering
    \graphicspath{ {./images/} }
    \includegraphics[scale=0.30,angle = 0]{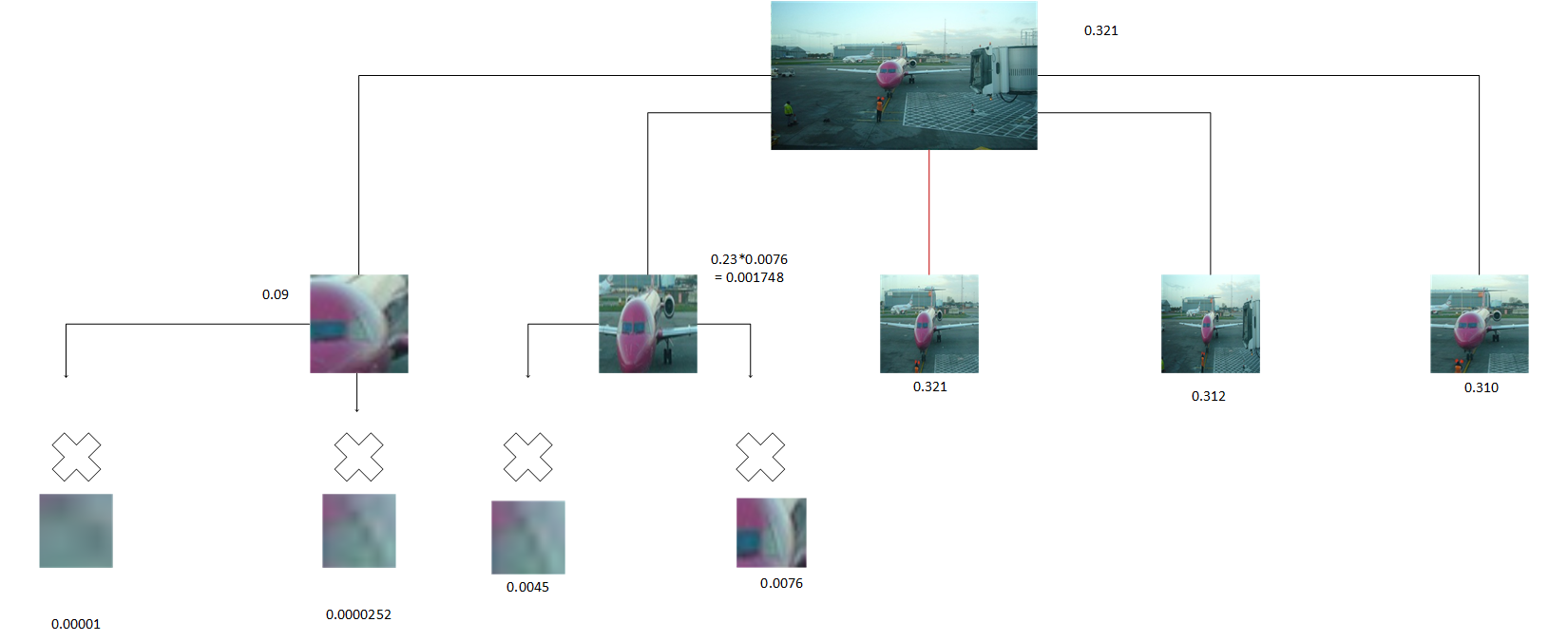}
    \caption{Object proposal tree pruning}
    \label{fig:Fig 6}
\end{figure}

\section{Experimental Results}

This section analyzes multiple aspects of both the segments discussed in the proposed work section i.e. the one – click method and the hierarchical object detection.

\subsection{One Click Annotation}

This section briefly discusses how our annotation approach differs from the standard annotation approach in the aspects of computational power, object scale, number of objects and different datasets. The feedback from users with little domain knowledge on deep learning and data annotation claimed the following for our approach of annotation:

\begin{itemize}
    \item Our approach of annotation saves incredible amount of time.
    \item Our approach is easier to use with when there are multiple objects to be annotated in the images.
    \item Our approach offers much less mental stress when objects are placed at a far depth from the point of capture as in the objects which are extremely small.
\end{itemize}

\subsubsection{Computational Power}
    
The table 1 shows the average time taken to annotate an image in GPU, CPU using our approach and the time taken to annotate a data using standard approach. The results clearly shows that our approach is advantageous as compared to the standard in case of both CPU and GPU as it 3 - 4 times faster as compared to the standard annotation process. Table 2 shows the time consumed by our approach with change in the type of GPU used. The results show that a decent GPU is enough to annotate the data with only a minute difference in the annotation time.

\begin{table}[!h]
\begin{center}
\begin{tabular}{ |c|c|c| }
\hline
Method & GPU & CPU \\ 
\hline
\textbf{Our Approach} & $\mathbf{20.8}$ & $\mathbf{33.6}$ \\ 
Standard Approach & 65.5 & 65.5 \\ 
\hline
\end{tabular}
\caption{Time comparison table (in seconds)}
\label{tab:Table 1}
\end{center}
\end{table}

\begin{table}[!h]
\begin{center}
\begin{tabular}{ |c|c|c|c| }
\hline
Method & \footnotesize{NVIDIA TITAN X}  & \footnotesize{GTX 1080 Ti} & \footnotesize{GTX 1050 Ti} \\ 
\hline
\textbf{\small{Our Approach}} & $\mathbf{20.8}$ & $\mathbf{22.5}$ & $\mathbf{24.9}$  \\ 
\hline
\small{Standard Approach} & 65.5 & 65.5 & 65.6 \\ 
\hline
\end{tabular}
\end{center}
\caption{Time comparison across multiple GPUs (in seconds)}
\label{tab:Table 2}
\end{table}

\subsubsection{Object Scale}

The table 3 shows the comparison between our approach and standard approach when the object size varies. The results in table 3 clearly indicate that our approach is better option when it comes to annotating object at smaller scale as compared to that of the standard approach of annotation, making it easier for the annotator by reducing the annotation time and mental load one has to carry around.

\begin{table}[!h]
\begin{center}
\begin{tabular}{ |c|c|c|c|c|c|c| }
\hline
\footnotesize{Method} & \footnotesize{300+} & $\footnotesize{300} < \footnotesize{200}$ & $\footnotesize{200} < \footnotesize{100}$ & $\footnotesize{100} < \footnotesize{50}$ & $\footnotesize{50} < \footnotesize{30}$ & $\footnotesize{30} < \footnotesize{20}$ \\
\hline
\textbf{\footnotesize{Our Approach}} & $\mathbf{\footnotesize{9.49}}$ &
$\mathbf{\footnotesize{9.39}}$ & $\mathbf{\footnotesize{7.9}}$ & $\mathbf{\footnotesize{7.48}}$ & $\mathbf{\footnotesize{7.51}}$ & $\mathbf{\footnotesize{6.15}}$ \\
\hline
\footnotesize{Standard Approach} & \footnotesize{14.1} & \footnotesize{12.2}  & \footnotesize{10.03}  & \footnotesize{9.07}  & \footnotesize{9.06}  & \footnotesize{6.55}  \\ 
\hline
\end{tabular}
\end{center}
\caption{Time comparison: size of the object (in seconds)}
\label{tab:Table 3}
\end{table}

\subsubsection{Number of Objects}

This section analyses the effectiveness of our approach when the number of objects in the images increase and the results for the same can be viewed in Table 4.The table 4 shows that the time difference increases radically when the number of objects in the image starts to increase.

\begin{table}[!h]
\begin{center}
\begin{tabular}{ |c|c|c|c|c|c| }
\hline
\footnotesize{Method} & \footnotesize{1} & \footnotesize{2} & \footnotesize{4} & \footnotesize{7}& \footnotesize{12+} \\ 
\hline
\textbf{\footnotesize{Our approach}} & \footnotesize{$\mathbf{5.15}$} & \footnotesize{$\mathbf{9.66}$} & \footnotesize{$\mathbf{21.4}$} &  \footnotesize{$\mathbf{32.01}$} & \footnotesize{$\mathbf{45}$} \\ 
\hline
\footnotesize{Traditional Approach} & \footnotesize{7.87} &  \footnotesize{15.66} &  \footnotesize{28.9} &  \footnotesize{40.7} &  \footnotesize{60.0} \\ 
\hline
\end{tabular}
\end{center}
\caption{Time comparison: number of objects (in seconds)}
\label{tab:Table 4}
\end{table}

\subsubsection{Different Datasets}
The our approach is applicable to all datasets such as PASCAL VOC \cite{everingham2010pascal}, KITTI \cite{geiger2013vision} and CITYSCAPES \cite{cordts2016cityscapes}. The table 5 shows the consistency of framework over multiple datasets proving that our approach is robust to changes in the distribution of the data.
\begin{table}[!h]
\begin{center}
\begin{tabular}{ |c|c|c|c| }
\hline
Method & PASCOL VOC & CITYSCAPES & KITTI \\ 
\hline
\footnotesize{\textbf{Our Approach}} & \footnotesize{$\mathbf{25.8}$}  & \footnotesize{$\mathbf{26.1}$}  & \footnotesize{$\mathbf{26.9}$}  \\ 
\hline
\footnotesize{Standard Approach} & \footnotesize{34.5}  & \footnotesize{52.8}  & \footnotesize{66.6}   \\ 
\hline
\end{tabular}
\end{center}
\caption{Time comparison: different data sets (in seconds)}
\label{tab:Table 5}
\end{table}

\subsection{Hierarchical Object Detection}

The hierarchical object detection also depends on multiple parameters influencing its accuracy and computational time, this section gives a brief overview about such parameters.

\subsubsection{Anchor-boxes}
The anchor boxes play a vital role in detecting the objects which were left out at the first iteration of detection but at the same time it takes a heavy toll on the computational time. The table 6 shows the trade off between accuracy and time based on the number of anchor boxes. The results show that with increase in number of anchor boxes the mean average precision score does increase but at the same time the computational time of annotation increases.
\\
\begin{table}[!h]
\begin{center}
\begin{tabular}{ |c|c|c|c| }

\hline
\footnotesize{Number of Anchors}
& \multicolumn{2}{|c|}{\footnotesize{Accuracy}} & \footnotesize{Time taken (seconds)}\\ 
\hline
$ $ & \footnotesize{Mean average precision} & \footnotesize{Recall} & $ $ \\
\hline
3 & 0.995 & 0.72 & 19.1  \\ 
\hline
5 & 0.997 & 0.801 & 20.8 \\ 
\hline
7 & 0.999 & 0.73 & 23.2 \\ 
\hline
\end{tabular}
\end{center}
\caption{Time vs accuracy comparison on varying number of anchors}
\label{tab:Table 6}
\end{table}
\subsubsection{Hierarchy count}
The hierarchy count is the number of iterations the model runs over the anchor box based object proposal, the table 7 discuss about the accuracy vs computational time trade-off. The results show with the increase size of hierarchy the results aren't much improving but only the time consumption increases.
\begin{table}[!h]
\begin{center}
\begin{tabular}{ |c|c|c|c| }
\hline
\footnotesize{Hierarchy count} & \multicolumn{2}{|c|}{\footnotesize{Accuracy}} & \footnotesize{Time taken (seconds)}\\ 
\hline
$ $ & \footnotesize{Mean average precision} & \footnotesize{Recall} & $ $ \\
\hline
\footnotesize{$2$} & \footnotesize{$0.997$} & \footnotesize{$0.805$} & \footnotesize{$19.5$}   \\ 
\hline
\footnotesize{$3$} & \footnotesize{$0.997$} & \footnotesize{$0.799$} & \footnotesize{$20.8$}  \\ 
\hline
\footnotesize{$4$} & \footnotesize{$0.997$} & \footnotesize{$0.797$} & \footnotesize{$22.2$}  \\ 
\hline
\footnotesize{$5$} & \footnotesize{$0.997$} & \footnotesize{$0.798$} & \footnotesize{$24.01$}  \\ 
\hline
\end{tabular}
\end{center}
\caption{Time vs accuracy comparison on hierarchy count}
\label{tab:Table 7}
\end{table}

\subsection{Detection Results}

\begin{table}[!h]
\begin{center}
\begin{tabular}{ |c|c|c|c|c| }
\hline
\footnotesize{Data set} & \multicolumn{2}{|c|}{\footnotesize{Hierarchical object detector}} & \multicolumn{2}{|c|}{\footnotesize{Standard object detector}}\\ 
\hline
$ $ & \footnotesize{Mean average precision} & \footnotesize{Recall} & \footnotesize{Mean average precision} & \footnotesize{Recall} \\
\hline
\footnotesize{PASCAL VOC} & \footnotesize{$\mathbf{0.999}$} & \footnotesize{$\mathbf{0.903}$} & \footnotesize{$0.85$} & \footnotesize{$0.451$}    \\ 
\hline
\footnotesize{KITTI} & \footnotesize{$\mathbf{0.997}$} & \footnotesize{$\mathbf{0.801}$} & \footnotesize{$0.912$} & \footnotesize{$0.61$}  \\ 
\hline
\footnotesize{CITYSCAPES} & \footnotesize{$\mathbf{0.924}$} & \footnotesize{$\mathbf{0.451}$} & \footnotesize{$0.227$} & \footnotesize{$0.09$}  \\ 
\hline
\end{tabular}
\end{center}
\caption{Comparison between results from standard object detector and hierarchical object detector}
\label{tab:Table 8}
\end{table}

The table 8 describes the performance of the hierarchical object detector on multiple datasets. It can be observed that the hierarchical object detector boosts the performance of annotation both in terms of mean average precision score as well as in recall score. The detection results for different types of scenarios are listed below where the detections from standard detector is compared to that of our approach.

\subsubsection{Correctly labelled and localized}
This is the case where the standard object detector is able to perform optimally by detecting all the object of interest in the image. The human annotated object center comes in handy here, the human annotated data containing the precise center co-ordinates are used instead of the network predicted centers to re-localize the detected objects. The changes in the looks of the bounding boxes can be viewed in Fig 7 and Fig 8. In Fig 8 we can observe that the objects are much more centered as compared to that of the Fig 7 the reason behind such loss turns out due to the loss of information occurring as a consequence of squeezing the image into a smaller grid losing some amount of information regarding the exact location of the object in order to obtain more dense feature for accurate classification of the object.
\begin{figure}[!h]
\begin{center}
\begin{tabular}{ c c c }
    \includegraphics[scale=0.6]{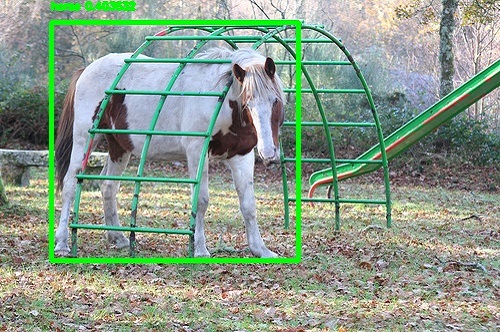}
& 
    \includegraphics[scale=0.3]{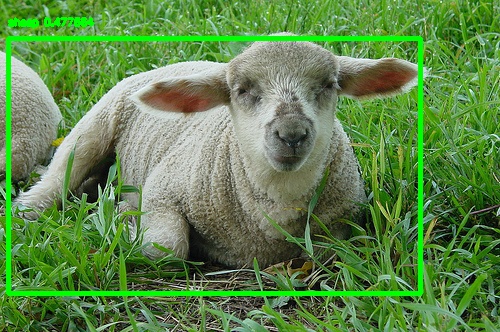}
 & 
    \includegraphics[scale=0.6]{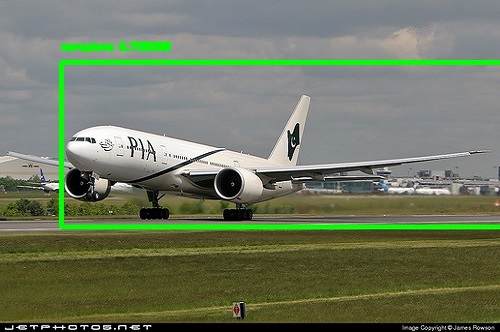}
\\  
\end{tabular}
\end{center}
\caption{Object detection results from the standard detector}
\label{fig:Figure 7}
\end{figure}

\begin{figure}[!h]
\begin{center}
\begin{tabular}{ c c c }
    \includegraphics[scale=0.6]{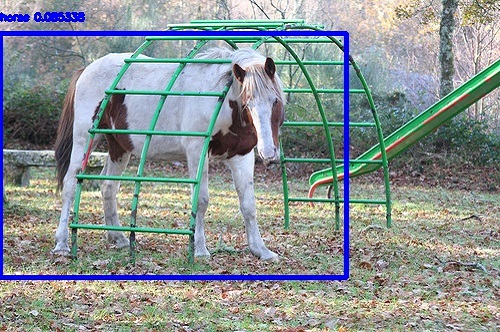}
 & 
    \includegraphics[scale=0.3]{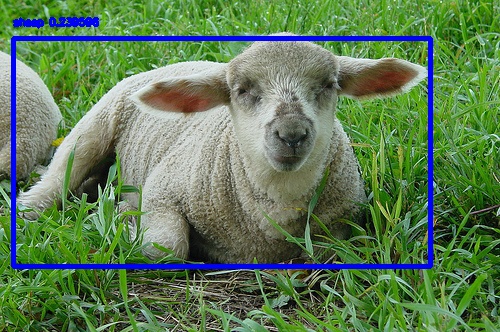}
 & 
    \includegraphics[scale=0.6]{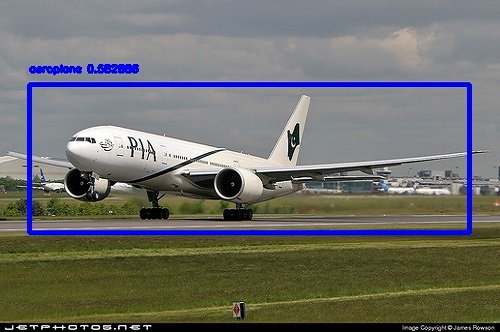}
 \\  
\end{tabular}
\end{center}
\caption{Improving correctly labelled and localized data using one - click annotation}
\label{fig:Figure 8}
\end{figure}

\subsubsection{Incorrectly labelled but correctly localized}
There are certain cases where the network will wrongly label the objects in the image which can be seen in Fig 9. The detections of such spurious nature are found by comparing the predicted object labels and human annotated object labels. The spurious detection are corrected using the label information and object center. The rest of the predicted information remains the same. The result of object detection after adding this detection can be seen in the Fig 10.

\begin{figure}[!h]
\begin{center}
\begin{tabular}{c c c}
    \includegraphics[scale=0.6]{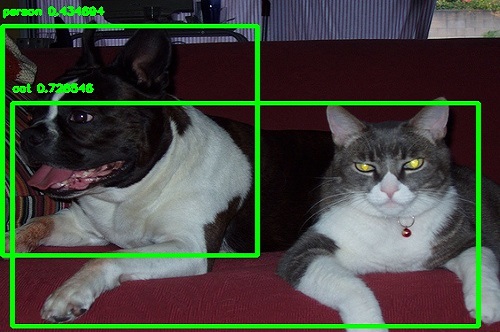}
& 
    \includegraphics[scale=0.6]{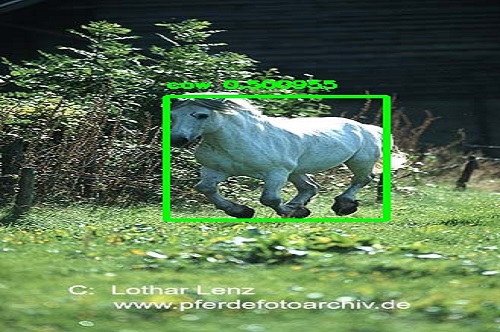}
 & 
    \includegraphics[scale=0.6]{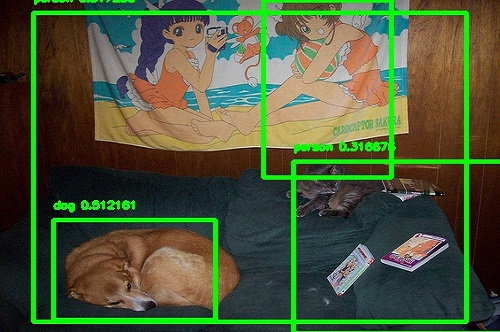}
\\  
\end{tabular}
\end{center}
\caption{Object detection results from the standard detector}
\label{fig: Figure 9}
\end{figure}

\begin{figure}[!h]
\begin{center}
\begin{tabular}{c c c}
    \includegraphics[scale=0.6]{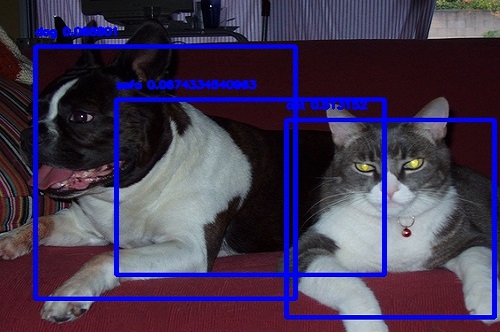}
 & 
    \includegraphics[scale=0.6]{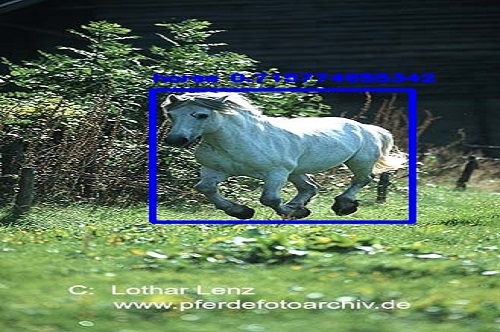}
 & 
    \includegraphics[scale=0.6]{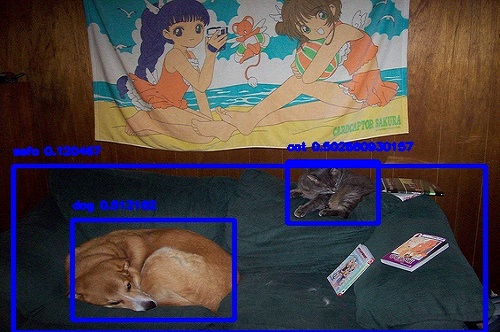}
 \\  
\end{tabular}
\end{center}
\caption{Improving incorrectly labelled but correctly localized data using one-click annotation}
\label{fig: Figure 10}
\end{figure}

\subsubsection{Incorrectly labelled and localized }
These set of detection are popularly termed as false positives. The false positives play an important role in determining the mean average precision of the object. The object detector possess a low mean average precision due to the incorrect classification and localization of the object in the image. Thus, for any annotated data it is desired to have very high mean average precision. The Fig 11 and 12 show that the results after removing such spurious detections.

\begin{figure}[!h]
\begin{center}
\begin{tabular}{c c c}   
    \includegraphics[scale=0.6]{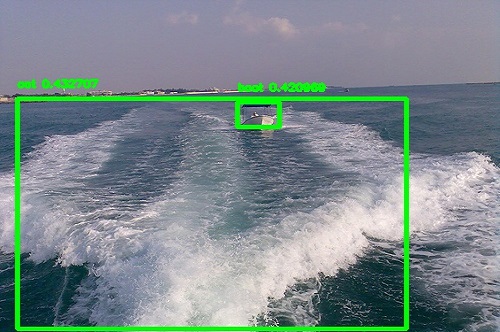}
& 
    \includegraphics[scale=0.6]{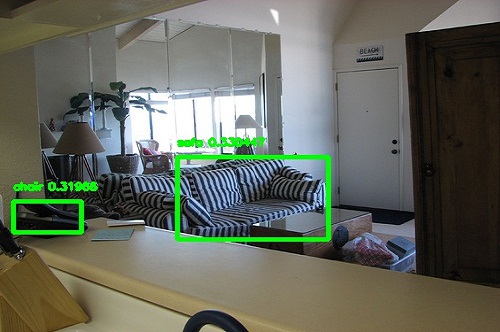}
 & 
    \includegraphics[scale=0.6]{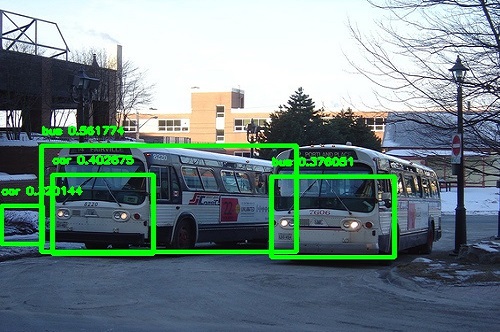}
 \\ 
\end{tabular}
\end{center}
\caption{Object detection results from a standard detector}
\label{fig: Figure 11}
\end{figure}

\begin{figure}[!h]
\begin{center}
\begin{tabular}{c c c}   
    \includegraphics[scale=0.6]{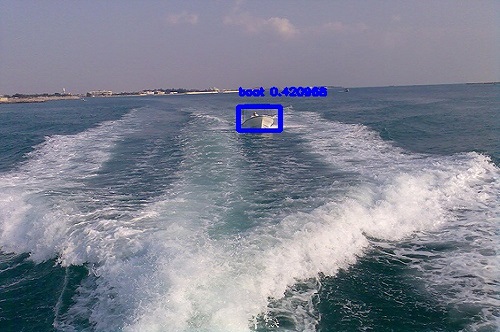}
 & 
    \includegraphics[scale=0.6]{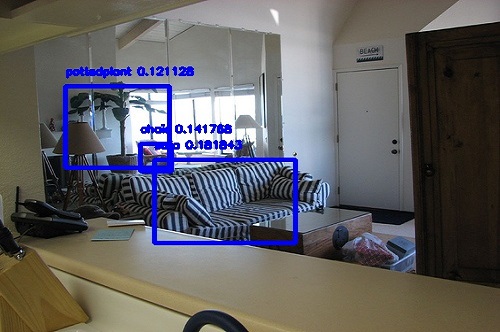}
 & 
    \includegraphics[scale=0.6]{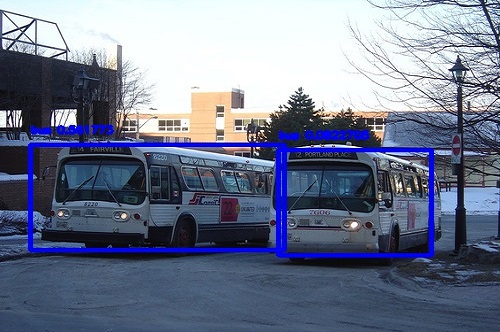}
 \\  
\end{tabular}
\end{center}
\caption{Improving incorrectly labelled and localized data using one-click annotation}
\label{fig: Figure 12}
\end{figure}

\subsubsection{Not labelled and localized}
In this subsection the case of missing out an object is explained. The hierarchical object detection comes into action in this region. The results for detection from standard object detection can be seen in Fig 13 and the results from hierarchical object detection can be seen in Fig 14.

\begin{figure}[!h]
\begin{center}
\begin{tabular}{c c c}
    \includegraphics[scale=0.6]{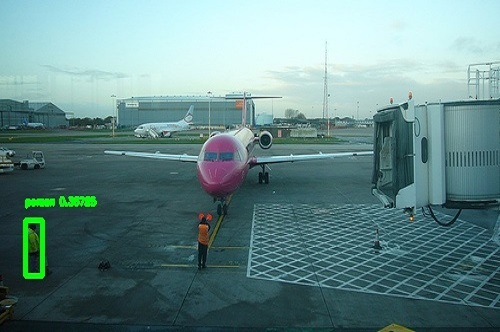}
& 
    \includegraphics[scale=0.6]{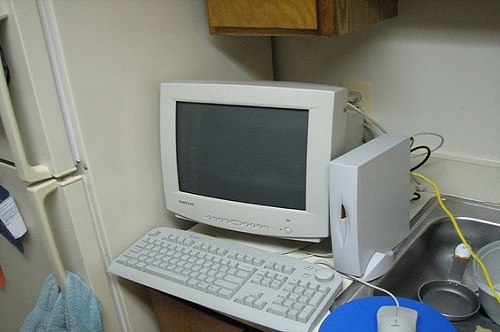}
 & 
    \includegraphics[scale=0.6]{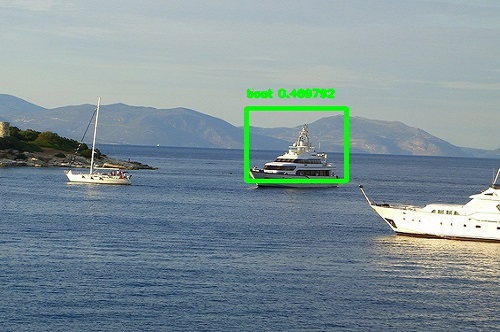}
 \\  
\end{tabular}
\end{center}
\caption{Object detection results from a standard object detector}
\label{fig: Figure 13}
\end{figure}

\begin{figure}[!h]
\begin{center}
\begin{tabular}{c c c}
    \includegraphics[scale=0.6]{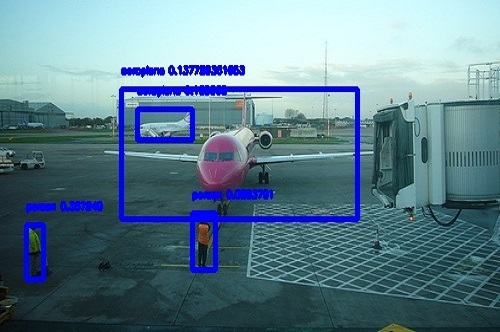}
 & 
    \includegraphics[scale=0.6]{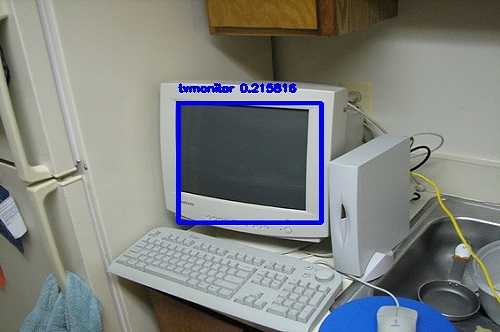}
 & 
    \includegraphics[scale=0.6]{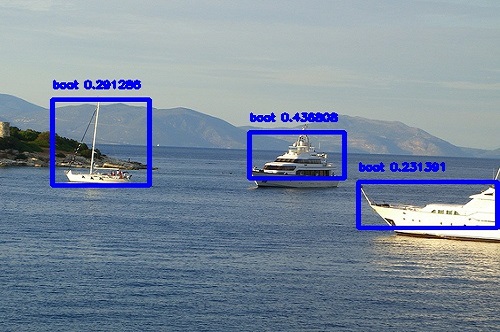}
 \\  
\end{tabular}
\end{center}
\caption{Hierarchical object detector on improving not labelled and localized data }
\label{fig: Figure 14}
\end{figure}
\section{Conclusion and Future work}
The proposed framework provides a novel solution to problem of annotating the data with least amount of human effort both mentally and physically by harnessing maximum amount of information with minimum amount of interaction with the computer. The current framework acts as the current state of the art object detector in case of guided object detection task by attaining a mean average precision score of 99.95 ,99.7,99.23 on PASCAL VOC, KITTI, CITYSCAPES and a recall score of 90.38,80.1,45.02 on PASCAL VOC, KITTI, CITYSCAPES respectively. The framework reduces the annotation time, cost and mental stress. The framework also removes the spurious human object center annotations reducing the time required for refining the annotation. The framework has proven to be 3 - 4 times faster than that of the standard annotation procedure. There lies a lot unexploited potential in the framework which can be further taken up for future research. One of them is that the network finds it difficult to annotate the similar objects which are placed together. In the first image in Fig 15 we can observe that although both the parrot were clicked by the annotator but one of the parrot masks its presence as it is prominent in comparison to the neighboring the object which was clicked, so a bounding box is created around the prominent one leaving out the clicked object center out of the box. This results only in a single detection.  The same trend can be observed in the other set of Figures 15 for the dogs in the second image and cows in third image.
\begin{figure}[!h]
\begin{center}
\begin{tabular}{c c c}
    \includegraphics[scale=0.6]{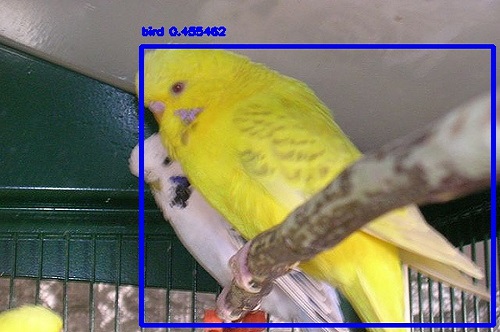}
& 
    \includegraphics[scale=0.6]{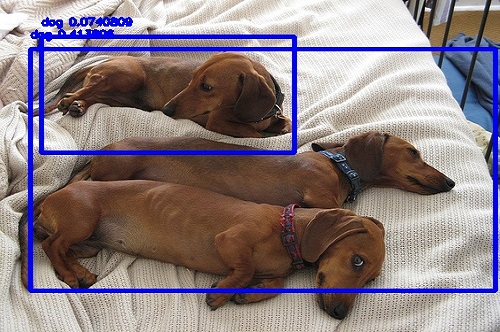}
 & 
    \includegraphics[scale=0.6]{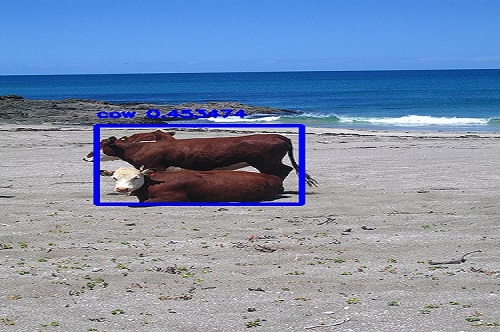}
 \\ 
\end{tabular}
\end{center}
\caption{Performance of hierarchical object detector when two similar objects are close}
\label{fig: Figure 15}
\end{figure}

\begin{figure}[!htp]
\begin{center}
\begin{tabular} {c c c}
    \includegraphics[scale=0.6]{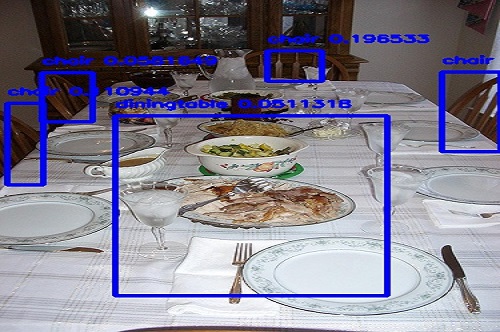}
& 
    \includegraphics[scale=0.6]{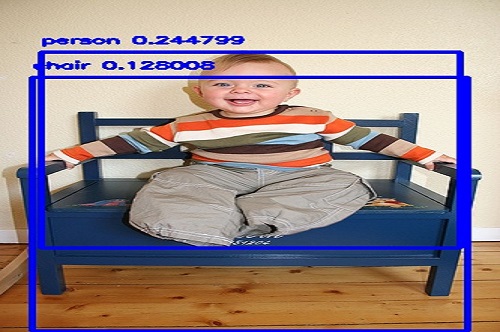}
 & 
    \includegraphics[scale=0.6]{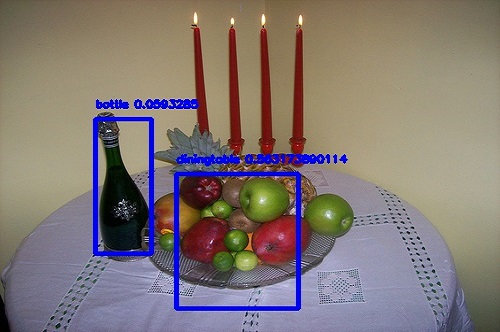}
 \\ 
\end{tabular}
\end{center}
\caption{Performance of hierarchical object detector when object centers are occluded}
\label{fig: Figure 16}
\end{figure}

Another future work lies in improving the object detection by allowing user to click any-where in the object as there are many situation where unwanted object might occlude the visual and spatial characteristics of the object of interest i.e an unwanted object might occlude the center point of the object of interest thus resulting in poor quality of anchor boxes to be used as an object proposal. An example for the same can be seen in Fig 16 where an object interest is occluded by an uninteresting object.

{\small
\bibliographystyle{ieee}
\bibliography{egbib}
}

\end{document}